\documentclass{ieeeaccess}
\usepackage{cite}
\usepackage{amsmath,amssymb,amsfonts}
\usepackage{algorithmic}
\usepackage{graphicx}
\usepackage{textcomp}
\usepackage{comment}

\usepackage{listings}
\usepackage{bm}
\makeatletter
\AtBeginDocument{\DeclareMathVersion{bold}
\SetSymbolFont{operators}{bold}{T1}{times}{b}{n}
\SetSymbolFont{NewLetters}{bold}{T1}{times}{b}{it}
\SetMathAlphabet{\mathrm}{bold}{T1}{times}{b}{n}
\SetMathAlphabet{\mathit}{bold}{T1}{times}{b}{it}
\SetMathAlphabet{\mathbf}{bold}{T1}{times}{b}{n}
\SetMathAlphabet{\mathtt}{bold}{OT1}{pcr}{b}{n}
\SetSymbolFont{symbols}{bold}{OMS}{cmsy}{b}{n}
\renewcommand\boldmath{\@nomath\boldmath\mathversion{bold}}}
\def\headerlogo{}
\def\headerlogoall{}
\def\ps@titlepage{%
  \def\@oddhead{}%
  \def\@evenhead{}%
  \def\@oddfoot{}%
  \def\@evenfoot{}%
}
\def\thevol{}
\def\theyear{}
\def\footervolfont{}
\providecommand{\doi}[1]{}
\providecommand{\history}[1]{}
\makeatother

\usepackage{etoolbox}
\makeatletter
\patchcmd{\@maketitle}
  {{\vss\doifont Digital Object Identifier\space\@doi\vss}\vspace*{7.4mm}\par}
  {}
  {}{}
\patchcmd{\@oddfoot}
  {{\footervolfont VOLUME\ \thevol, \theyear}}
  {}
  {}{}
\patchcmd{\@evenfoot}
  {{\footervolfont VOLUME\ \thevol, \theyear}}
  {}
  {}{}
\makeatother

\def\BibTeX{{\rm B\kern-.05em{\sc i\kern-.025em b}\kern-.08em
    T\kern-.1667em\lower.7ex\hbox{E}\kern-.125emX}}

\begin{document}

\title{Free Energy–Based Modeling of Emotional Dynamics in Video Advertisements}
\author{\uppercase{Takashi Ushio}\authorrefmark{1}\authorrefmark{*}, \uppercase{Kazuhiro Onishi}\authorrefmark{2},
and \uppercase{Hideyoshi Yanagisawa}\authorrefmark{3}}

\address[1]{Hakuhodo DY Holdings Inc., Tokyo, Japan (e-mail: takashi.ushio@hakuhodo.co.jp)}
\address[2]{Hakuhodo Technologies Inc., Tokyo, Japan (e-mail: kazuhiro.oonishi@hakuhodo-technologies.co.jp)}
\address[3]{Graduate School of Engineering, The University of Tokyo, Tokyo, Japan (e-mail: hide@mech.t.u-tokyo.ac.jp)}
\tfootnote{
This article has been accepted for publication in IEEE Access and will be published shortly.
This research did not receive any specific grant from funding agencies in the public, commercial, or not-for-profit sectors.}


\corresp{\authorrefmark{*}Corresponding author: Takashi Ushio (e-mail: ushio.takashi.jp@gmail.com).}

\begin{abstract}
Emotional responses during advertising video viewing are recognized as essential for understanding media effects because they have influenced attention, memory, and purchase intention. To establish a methodological basis for explainable emotion estimation without relying on external information such as physiological signals or subjective ratings, we have quantified “pleasantness,” “surprise,” and “habituation” solely from scene-level expression features of advertising videos, drawing on the free energy(FE) principle, which has provided a unified account of perception, learning, and behavior. In this framework, Kullback–Leibler divergence (KLD) has captured prediction error, Bayesian surprise (BS) has captured belief updates, and uncertainty (UN) has reflected prior ambiguity, and together they have formed the core components of FE. Using 1,059 15 s food video advertisements, the experiments have shown that KLD has reflected “pleasantness” associated with brand presentation, BS has captured “surprise” arising from informational complexity, and UN has reflected “surprise” driven by uncertainty in element types and spatial arrangements, as well as by the variability and quantity of presented elements. This study also identified three characteristic emotional patterns, namely uncertain stimulus, sustained high emotion, and momentary peak and decay, demonstrating the usefulness of the proposed method. Robustness across nine hyperparameter settings and generalization tests with six types of Japanese advertising videos (three genres and two durations) confirmed that these tendencies remained stable. This work can be extended by integrating a wider range of expression elements and validating the approach through subjective ratings, ultimately guiding the development of technologies that can support the creation of more engaging advertising videos.
\end{abstract}


\titlepgskip=-21pt

\maketitle

\section{Introduction}\label{sec:intro}
In recent years, understanding the emotional responses of viewers to video advertisements has become increasingly important for improving user experience, enhancing advertising effectiveness, and supporting creative design. Emotions such as pleasantness, surprise, and interest strongly influence purchase intention and brand memory, which makes it essential to quantify them both during the production stage and in post-hoc evaluations. Video advertisements consist of multiple scenes and diverse expressive elements that creators intentionally combine. As a result, estimating the emotions of viewers in a manner that is both accurate and interpretable remains challenging.

Physiology-based approaches have been widely used to estimate emotional responses, using facial expressions, heart rate, electrodermal activity, and brain activity \cite{Otamendi2020-ac, McDuff2015-sm, Vecchiato2011-bg, Caruelle2024-dz}. Although they are effective for predicting advertisement liking and purchase intention, these approaches are sensitive to environmental noise and individual differences. Content-based deep learning models that rely on audio and visual features \cite{Antonov2024-ym, Yang2018-qk, Yang2021-yk} have also shown strong performance. However, they require large quantities of annotated data, and additional data preparation is needed for different durations, cultures, and genres. As these models are black boxes, it is difficult to explain which expressive elements contribute to the estimated emotional responses. In parallel with these approaches, the free energy principle(FEP), predictive processing, and active inference have provided a unified theoretical account of perception, learning, and emotion in terms of minimizing free energy(FE) or the prediction error\cite{Friston2006-hi, Friston2010-xn}. Previous studies have reported that fluctuations in FE correspond to positive and negative valence \cite{Joffily2013-sm}, and that Bayesian surprise influences attention and emotion \cite{Itti2009-id, Baldi2010-qv, Yanagisawa2019-jq}. However, most applications of these theories to real video stimuli have focused on video quality or low-level visual processing. A systematic framework that uses the FEP to analyze expressive structures in real advertising videos at the level of scenes and expressive elements has not yet been established.

The aim of this study is to propose a methodology that systematically explains how expressive elements in video advertisements induce emotional responses without relying on external information, such as physiological data or subjective ratings. The proposed method divides videos into scenes and uses the temporal expressive features of each scene as inputs to the generative model. Based on the FEP, this study quantifies pleasantness, surprise, and habituation using FE-related indices computed from the model.
A key feature of this study is that it uses only visual and auditory information to provide a mathematical explanation of the relationship between expressive elements and emotional indices. This allows the visualization of the expressive patterns that evoke specific emotional responses. This study also shows that the proposed scene- and video-level indices are robust to variations in model parameters and sampling. Furthermore, it is demonstrated that the framework generalizes to a wide range of advertising videos that differ in genre and duration.
Overall, this study provides a unified computational framework for understanding how temporal advertising expressions shape emotional dynamics across diverse video contexts.

\section{Background}\label{sec:bg}
\subsection{FEP}\label{sec:foundations_fe}

The FEP is a theoretical framework proposed by Friston \cite{Friston2006-hi, Friston2010-xn}. It states that perception, learning, and action in living organisms are determined to minimize surprise, expressed as FE, and this process enables organisms to adapt to their environment. Active inference, which derives the FEP as a first principle, has proposed two methods for reducing the mismatch between the internal model of an agent and the external world \cite{Parr2022-pl}. 

The first method involves changing the internal states to better match the world through perception and learning, which minimizes variational free energy(VFE). The second method involves changing the world to better match the internal model through action. It minimizes the expected FE by evaluating each possible action based on its anticipated future outcomes. Focusing on perception and learning, VFE is expressed as follows \cite{Friston2006-hi, Friston2010-xn,Parr2022-pl}. 

\begin{equation}\label{eq:parception_fe}
F(q, o_\tau) = D_{\mathrm{KL}}\left[q(s_\tau) \,\|\, p(s_\tau \mid o_\tau)\right] - \ln p(o_\tau)
\end{equation}

where $q(s_{\tau})$ and $p(s_{\tau}\mid o_{\tau})$ represent the prior distribution before observation and the posterior distribution derived from the generative model after observation. The first term captures the amount of information gained from the prior to the posterior, expressed as the Kullback Leibler divergence(KLD), and reflects the reduction in FE during perception. The second term represents the Shannon surprise(S) of the observation. Minimizing FE means predicting $q$ so that it approaches $p$. In practice, the state distribution $q(s_{\tau})$ can be obtained using standard inference schemes such as marginal message passing \cite{Parr2019-dx}.

In the learning process, FE is defined as shown in (\ref{eq:learning_fe}) \cite{Friston2006-hi, Friston2010-xn,Parr2022-pl}, and the previously described S is minimized.

\begin{equation}\label{eq:learning_fe}
F(q, o_\tau) = D_{\mathrm{KL}}\left[q'(s_\tau) \,\|\, p(s_\tau)\right] - \mathbb{E}_{q'(s_\tau)} \left[ \ln p(o_\tau \mid s_\tau) \right]
\end{equation}

The first term represents the amount of information gained through learning, which corresponds to Bayesian surprise (BS), as well as the reduction in FE that accompanies the update of the prior distribution. The second term represents the uncertainty of recognition (UN). Equations (\ref{eq:parception_fe}) and (\ref{eq:learning_fe}) can be computed by transforming the Kullback Leibler divergence between $p(s,o)$ and $q(s)$ based on Bayes theorem, and they allow different interpretations of the same quantity. 

Assuming a two-stage process of perception and learning, learning replaces $q$ with the post-perception state distribution $q^{\prime}(s_{\tau})$ and updates $p$ so that it approaches $q^{\prime}$ \cite{Yanagisawa2023-uq}.

\subsection{Relationship between Free Energy and Emotion}

Many researchers have discussed the relationship between the FEP and emotion. Regarding subjective pleasantness, Joffily et al.\cite{Joffily2013-sm} have related decreases and increases in FE over time to positive and negative emotional valence, respectively. Yanagisawa et al.\cite{Yanagisawa2023-uq} modeled the transition from unconscious perception to conscious learning as a switch in the Bayesian prior distribution, and showed that changes in FE during perception and learning correspond to emotions such as pleasantness, interest, confusion, and boredom.

Regarding subjective surprise, S in information theory represents how unexpected an observation $o_{\tau}$ is, and it is defined as Shannon surprise, expressed as S. This quantity is directly related to entropy, which represents the average level of surprise across all possible observations. Entropy is the expectation of S, and higher entropy indicates a greater average level of surprise associated with observations. Within the framework of the FEP, S appears explicitly as the second term of the VFE in (\ref{eq:learning_fe}), representing the magnitude of surprise processed through learning. Therefore, the components of S, BS, and UN naturally correspond to subjective surprise.

Yanagisawa et al.\cite{Yanagisawa2019-jq} conducted theoretical and empirical studies on the relationship between Bayesian surprise in (\ref{eq:learning_fe}) and subjective feelings of surprise. Yanagisawa\cite{Yanagisawa2021-sq} also proposed a mathematical model that explains arousal potential based on novelty, complexity, and uncertainty. In addition, the sum of the KLD and BS has shown an inverted U-shaped relationship with S\cite{Yanagisawa2025-rs}.

FE is also related to the cognitive load required to process a stimulus. A state of high FE involves large prediction errors and high uncertainty, and therefore requires more information processing and model updating, which can lead to a higher cognitive load. In contrast, when an organism adapts to a stimulus and FE decreases, the prediction errors that must be processed become smaller, and the cognitive burden is reduced. Ueda et al.\cite{Ueda2021-hb} showed that repeated exposure to the same stimulus gradually weakens emotional responses and that the FEP can account for this habituation.

\subsection{Applicability of the FEP to Multimodal Media}
Empirical studies support the FEP-based accounts of emotions. For example, Itti et al. \cite{Itti2009-id, Baldi2010-qv} showed that BS computed from visual features, such as luminance and motion in video scenes, corresponds to subjective surprise and drives visual attention. Usuda et al. \cite{Usuda2022-zj} quantified the KLD as pleasure and S as surprise in response to music stimuli. 

Other studies described aesthetic pleasure and affective valence as outcomes of fluctuations in prediction errors \cite{Sarasso2022-rw, Frascaroli2024-ja}. Honda et al. \cite{Honda2025-zq} suggested that embedding an optimal level of S in the behavior of a product (a virtual wristwatch) maximized user interest. Ueda et al. \cite{Ueda2025-fa} investigated the neural mechanisms of the emotion of “interest” that arise when recognizing new information from a different perspective using a functional magnetic resonance imaging (fMRI) experiment based on the FE model. Researchers have applied the FEP to video quality assessment and have shown how video quality and editing techniques influence the viewing experience \cite{Xu2016-la, Wang2025-ju}.

The FEP has the potential to explain emotional responses that people exhibit toward a wide range of stimuli. However, no study has applied this principle to estimate emotion in advertising videos that contain multiple deliberately constructed scenes and expression elements. If the influence of higher-level expression elements on viewer emotions can be clarified and design patterns can be identified from these relationships, such knowledge could inform guidelines for video production and support post-production evaluation. Ultimately, this may contribute to video designs that enhance emotional valence and purchase intention.

\section{Methodology}\label{sec:methodology}

\begin{figure}[t]
\centering
\includegraphics[width=\columnwidth]{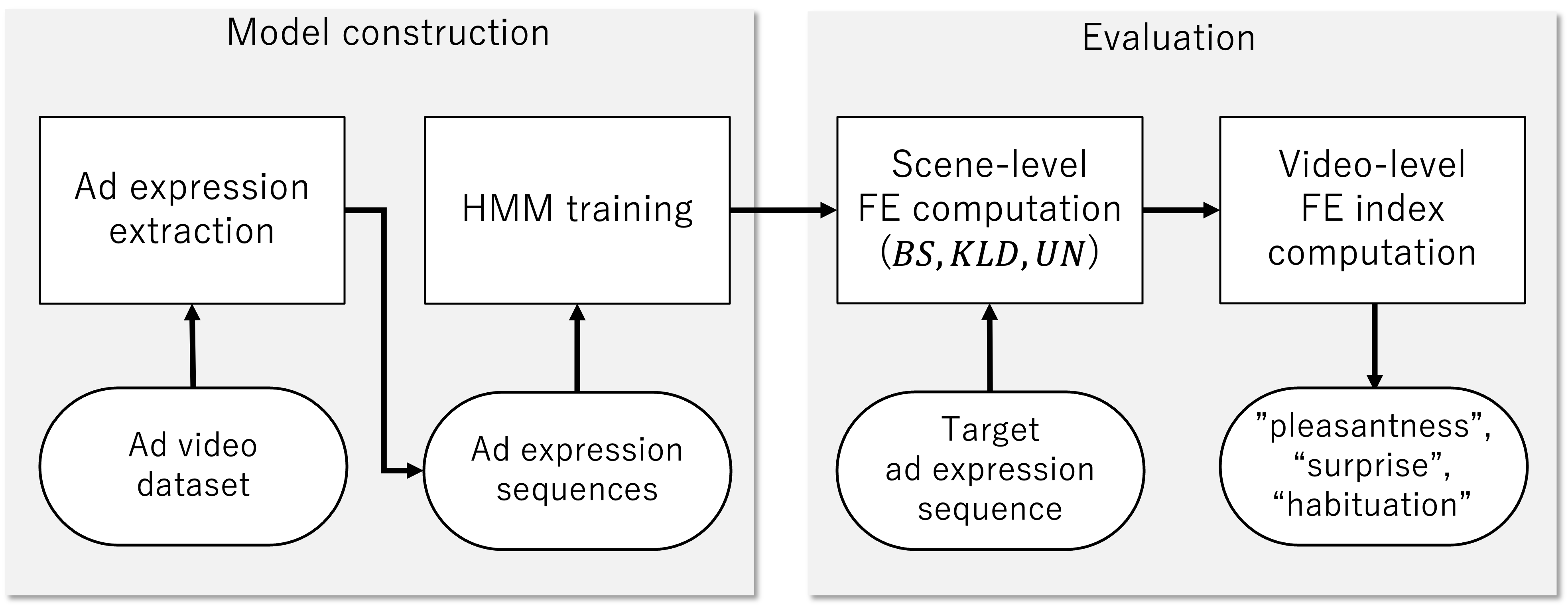}
\caption{Overview of the proposed processing flow.}
\label{fig:video_fep_overview}
\end{figure}

This study develops a method that estimates emotional responses using multi-modal expressive elements extracted from video advertisements and a generative model based on the FEP. The proposed method consists of two processes: the model construction process and the evaluation process (Fig.~\ref{fig:video_fep_overview}).

In the model construction process, vision-language models (VLMs) and large-language models (LLMs) are used to automatically extract expressive elements from an advertising video dataset. These elements are then structured into scene-level sequences. Using these sequences as observed variables, a categorical hidden Markov model (HMM) is trained to capture the latent states that arise during viewing and the transitions among these states.

In the evaluation process, FE is computed for each scene using the trained HMM and the structured expressive sequences. Specifically, three components of FE are derived based on the mismatch between the HMM predictions and the observations. These components are the KLD, which represents information gain, BS, which reflects the complexity of the stimulus, and UN, which represents recognition uncertainty. These scene-level indices are then integrated into video-level indices by computing summary statistics, such as the maximum value, the final value, skewness, and the decay rate. Based on prior studies\cite{Joffily2013-sm, Yanagisawa2023-uq, Yanagisawa2019-jq, Yanagisawa2021-sq,Yanagisawa2025-rs, Ueda2021-hb}, KLD is used as an index of pleasantness and BS and UN are used as indices of surprise. In addition, habituation is defined as the reduction in these indices under repeated exposure.

This methodology analyzes how temporally varying expressive elements in advertisements evoke emotional responses in viewers, and how these responses evolve when repeated viewing is simulated using the same stimulus. The proposed approach provides a unified and interpretable framework for understanding emotional dynamics in video advertisements.

\section{Proposed Algorithms}
Based on the methodological framework introduced in Section~\ref{sec:methodology}, this section presents the specific algorithms used to implement the proposed approach.

\subsection{Advertising Video Element Taxonomy}\label{sec:ad_element}
The structure of video advertisements depends on historical context and the tacit knowledge of creators. Thus far, numerous companies have developed design patterns as systematic frameworks to address this issue and improve advertising effectiveness (e.g., reach and attitude change). One framework that recently gained considerable attention is the ABCD framework proposed by Google \cite{googleABCD}. This framework encourages using expressions that serve the four expected functions for viewers: attract (gain attention), brand (increase brand recognition), connect (promote empathy), and direct (drive action). Studies showed that these elements are related to advertising effectiveness.

Specific expressive elements that correspond to the ABCD framework are diverse, and it is not realistic to handle them comprehensively. Therefore, this study defines “advertising expression elements” as visual and auditory features that can be mechanically extracted and judged in the binary form (0/1) for their presence or absence. Furthermore, the correspondence between these elements and the ABCD framework were organized as listed in Table~\ref{tab:ad_elements}.

\begin{table}[t]
\centering
\caption{\textbf{Advertising Expression Elements.}}
\label{tab:ad_elements}
\renewcommand{\arraystretch}{1.1}
\begin{tabular}{p{0.18\columnwidth} p{0.35\columnwidth} p{0.35\columnwidth}}
\hline
\textbf{Message type} & \textbf{Visual expression} & \textbf{Auditory expression} \\
\hline
Attract & Product close-up and character close-up 
        & Catchphrase \\ \hline
Brand   & Logo and product image 
        & Brand, company, and product name \\ \hline
Connect & Characters 
        & Addressing and positive words \\ \hline
Direct  & Product description, call to action, and motivational message 
        & Product description, call to action, and motivational message \\
\hline
\end{tabular}
\end{table}

\subsection{Multimodal Feature Extraction with LLM and VLM}
The method for automatically extracting advertising expression elements from visual and auditory modalities of video advertisements includes extracting visual and auditory expression elements.

The visual expression elements were extracted in three steps using a VLM: scene segmentation, representative frame selection, and classification. To this end, the video was segmented into scenes based on frame changes using PySceneDetect \cite{pyscenedetect} and the middle frame of each scene was extracted as its representative still image. Prompts for advertisement expression classification were input into MiniCPMV2.6 \cite{minicpmv}, which is a type of VLM, and the presence or absence (0/1) of each advertising expression element listed in Table~\ref{tab:ad_elements} was determined. Expressions were merged when the same expression continued across adjacent scenes.

Demucs \cite{demucs} was used for source separation to isolate the voice channel and extract the auditory expression elements. The speech was transcribed using Whisper \cite{whisper}. Finally, the presence or absence (0/1) of each advertising expression element listed in Table~\ref{tab:ad_elements} was determined using phi-4 \cite{phi-4}, which is an LLM. Each utterance was aligned with the closest video scene and integrated with visual expression elements because Whisper outputs the start and end times of each utterance.

Thus, multi-modal advertising expression elements were integrated at the scene-level while maintaining temporal consistency, and they were used as inputs to construct the subsequent generative model. The details of the prompts for the VLM and LLM are provided in Appendix A~\ref{appendix:prompt}.

\subsection{HMM-Based Generative Modeling}
This study considered video advertisements as stimuli and built a generative model for simulating the perceptions of and learning processes of viewers based on the FEP. The model describes how sequences of observations in the visual modality $\mathbf{o}^1=\{o_1^1,o_2^1,\ldots,o_{\tau}^1\}$ and auditory modality $\mathbf{o}^2=\{o_1^2,o_2^2,\ldots,o_{\tau}^2\}$ are generated from a sequence of hidden states $\mathbf{s}=\{s_1,s_2,\ldots,s_{\tau}\}$. Here, $s_\tau$ represents the hidden state at time $\tau$ (e.g., a semantic or emotional internal state) and $o_\tau$ represents the advertising expression elements observed at that time. The structures of states were not in advance, and instead, they were estimated from the observed data. Observations were separated by modality and defined as combinations of visual and auditory advertising expression elements (e.g., “Logo \& Characters” and “Catchphrase \& Brand”).

\begin{figure}[t]
\centering
\includegraphics[width=\columnwidth]{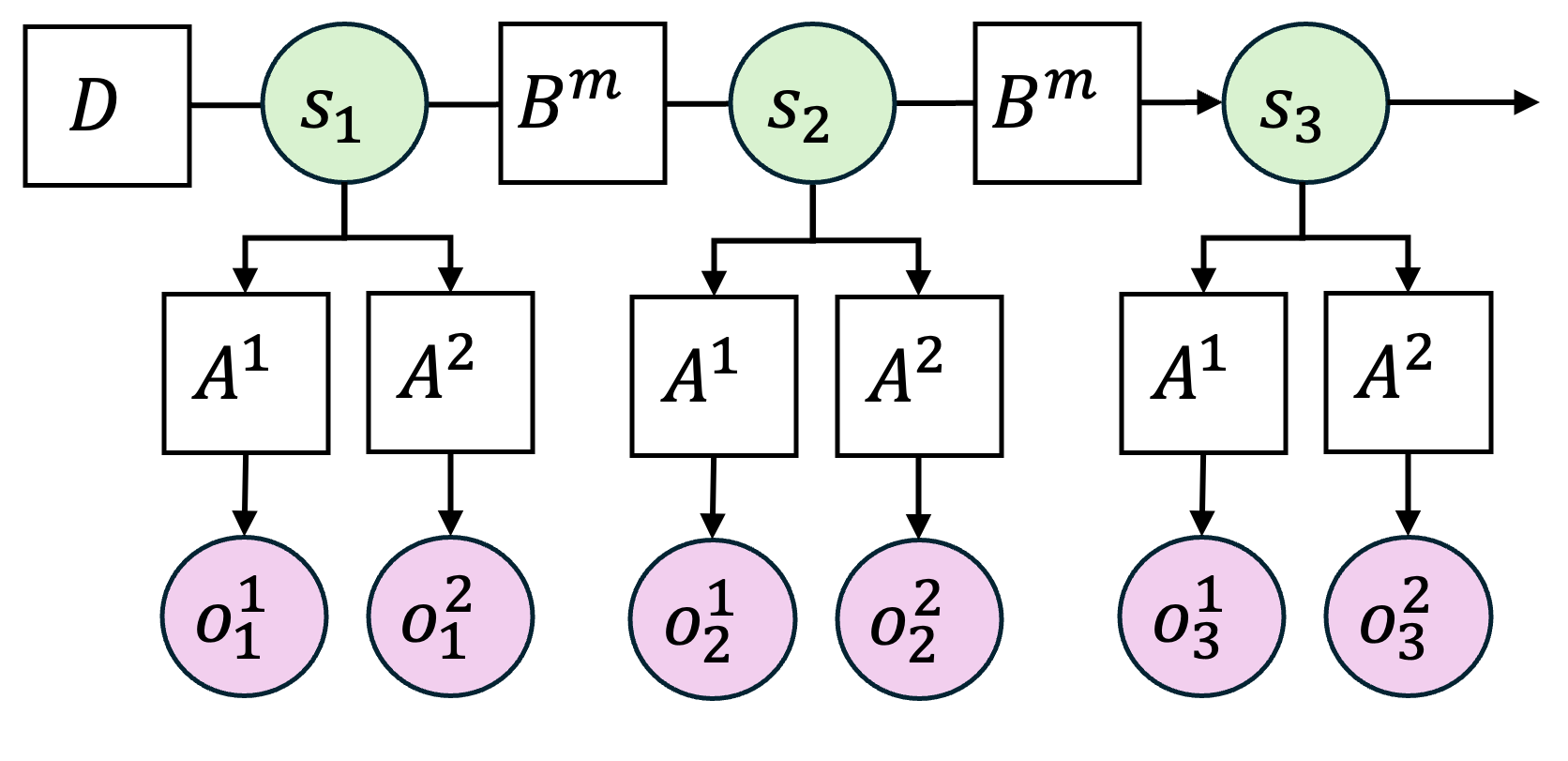}
\caption{Graphical model of the proposed HMM.}
\label{fig:graphical_model}
\end{figure}

The generative model was formulated as an HMM. The visual observation probabilities $\mathbf{A}^1$, auditory observation probabilities $\mathbf{A}^2$, state transition probabilities $\mathbf{B}^m$, and initial belief $\mathbf{D}$ were represented using categorical distributions. Each parameter had a Dirichlet prior that enabled a flexible representation. Furthermore, separate transition matrices $\mathbf{B}^m$ were introduced for fixed 3 s segments, dividing each video accordingly (e.g., 15 s into 5 segments and 30 s into 10 segments). This transition was used to represent state transitions based on the position in the video, and the effect of inserting advertising expression elements was captured. Figure~\ref{fig:graphical_model} illustrates the overall structure of the generative model.

\subsection{Scene-Level FE Computation}\label{sec:scene_fpe}
For the constructed generative model, when the observation of each scene $o_\tau$ was entered, the KLD was computed, which reflects the information gain from the prior to the posterior (Section~\ref{sec:foundations_fe}). The post-perception state distribution $q'(s_\tau)$ was estimated using marginal message passing, a commonly used scheme for FE minimization \cite{Parr2019-dx}.

During the learning process, BS and UN were computed, which constitute additional components of the FE, using the updated state distribution $q'(s_\tau)$.

\subsection{Video-Level Emotional Index Computation}\label{sec:video_fpe}
As video advertisements have multiple scenes, the FE metrics (KLD, BS, and UN) calculated for each scene must be integrated to derive representative emotional indices for the entire video. In this study, the characteristics of “pleasantness” and “surprise,” as well as their “habituation” were extracted at the video-level by summarizing temporal changes in these indices.

The representative values of each FE metric were computed based on the perspectives described later, which enabled the description of the emotional characteristics of the entire video.

\subsubsection{Peak-End Effect}\label{sec:peak_end}
According to studies by Fredrickson and Kahneman \cite{Fredrickson2000-mf, Kahneman2000-vb}, people tend to evaluate experiences not by their average but by their salient moments, particularly the emotional peak or end. This “peak–end effect” applies to video viewing, where certain scenes may strongly affect the overall emotional evaluation.

In this study, the peak (maximum value) and end (final scene) of FE metrics were used as representative values of emotional characteristics for each video. Let $\mathcal{C} $ represent the set of scenes contained in a video with the total number of scenes $ N = |\mathcal{C}| $. For each scene $c \in \mathcal{C}$, FE metrics were calculated as $ M(c) \in \{ \text{BS}(c), \text{KLD}(c), \text{UN}(c) \} $. Based on this, the peak value of $ \mathrm{peak}^{(M)} $ and end value of $ \mathrm{end}^{(M)} $ were computed.

These values quantified the emotional intensity of scenes that create a strong impression on viewers and served as emotional indices at the video-level.
\subsubsection{Distribution of Emotional Responses}
Parducci \cite{Parducci1965-xh, Parducci1995-kt} found that people evaluate stimuli in experiences relative to other items in a set and rely heavily on extreme values (upper and lower bounds) and their frequency of occurrence when forming evaluations. Based on this range–frequency theory \cite{Smith1989-ts}, the skewness of emotional response distributions was analyzed, which showed that negatively skewed distributions were associated with more positive evaluations.

In this study, the skewness of the FE metrics $\mathrm{Skew}^{(M)}$ was derived for each scene to capture the shape of the emotional response distribution. Skewness allows us to consider whether and to what extent the distribution is biased toward higher or lower values, thereby characterizing the asymmetry of emotional reactions. As a single peak or average value cannot describe the overall response pattern, this asymmetry can be interpreted as a characteristic of viewers’ emotional tendencies.

\subsubsection{Habituation to Stimuli}
Viewers often watch videos repeatedly, and the resulting reduction in responses was modeled as “habituation.”
In this study, repeated viewing was simulated $r$ times, and the set of HMM parameters $\Theta$ was updated at each repetition to quantify this decay. Then, the decay rate for the FE index $M$ is defined as follows:

\begin{equation}\label{eq:decayrate}
\mathrm{Decay \space rate}^{(M)} =
\frac{\sum_{c \in \mathcal{C}} \left[ M^{(1)}(c) - M^{(r)}(c) \right]}
{\sum_{c \in \mathcal{C}} M^{(1)}(c)}
\end{equation}

where $M^{(1)}(c)$ and $M^{(r)}(c)$ represent the value of the index for scene $c$ at the first viewing and the value at the rth viewing, respectively. This ratio indicates the extent to which emotional responses elicited by a video diminish with repeated viewing, and it is used as an index for evaluating the sustained effect of advertising expressions under repeated exposure.

\subsection{Implementation Details and Experimental Settings}\label{sec:fe_feature}
Visual and audio-based expression elements were extracted from each scene in the video advertisement dataset, and a time-series dataset of advertising expressions was constructed. The observed variables were then separated by modality using combinations of visual and audio expression elements as inputs.

The number of scenes per video, the types of visual and audio expression elements, and the maximum sequence length vary depending on the dataset used. 

In the generative model, Dirichlet priors were assigned to the observation probability matrices $\mathbf{A}^1$ and $\mathbf{A}^2$, 
the state transition matrix $\mathbf{B}$, and the initial state 
distribution $\mathbf{D}$. The set of parameters governing these Dirichlet distributions is denoted by $\boldsymbol{\Theta}$. Using variational inference implemented in Pyro, the probability matrices themselves were not optimized. Instead, the 
variational posterior distributions were also defined within the 
Dirichlet family, and the concentration parameters in 
$\boldsymbol{\Theta}$ were learned directly from the data. 
In this way, the expected forms of $\mathbf{A}^1$, $\mathbf{A}^2$, 
$\mathbf{B}$, and $\mathbf{D}$ were inferred implicitly through the learned variational parameters.

Additionally, the mini-batch learning rate, the Dirichlet prior scale, and the number of hidden states were tuned. The optimal configuration (learning rate = 0.0275, Dirichlet scale = 0.2, and five hidden states) was selected based on the validation evidence lower-bound (ELBO). Further details are provided in Appendix B~\ref{appendix:hyperparam}.

Based on the trained model, scene-level FE metrics $M(c) \in \{\mathrm{BS}(c), \mathrm{KLD}(c), \mathrm{UN}(c)\}$ were computed using the methods defined in Section~\ref{sec:scene_fpe}. The representative video-level indices, including $\mathrm{peak}^{(M)}$, $\mathrm{end}^{(M)}$, $\mathrm{Skew}^{(M)}$, and $\mathrm{Decay Rate}^{(M)}$ in Section~\ref{sec:video_fpe} were then derived.

To compute $\mathrm{Decay Rate}^{(M)}$, repeated viewing was simulated by inputting the same observation sequence five times. After each simulated viewing, $\boldsymbol{\Theta}$ was updated using a learning rate of 10.0. This learning rate applied only to the single sample updates performed during the repeated-viewing simulation, not to mini-batch training. All implementations were performed in Python.

\section{Main Results}
The main question addressed by this study is: "How can the emotional effects of video advertisements, including pleasantness, surprise, and habituation, be mathematically quantified using expression elements and the FEP?" This section presents empirical results that answer this question and summarizes the main contributions of the proposed framework.

\subsection{Video Advertisement Dataset}
The empirical analysis used television video advertisements from three industries (food, cosmetics, and automotive) in Japan between January 2022 and July 2024, with two durations (15 and 30 s). The number of samples and the detected number of scenes for each dataset are listed in Table~\ref{tab:dataset_stats}.
In addition to sample size, the datasets differ in the variety of detected visual and auditory expression-element types and in the maximum sequence length, reflecting differences in production styles, narrative tempo, and audio design across categories.

\begin{table}[t]
\centering
\caption{Video Ad Dataset Overview}
\label{tab:dataset_stats}
\begin{tabular}{llllll}
\hline
Dataset 
& Videos & Scenes 
& Vis. & Aud. & Len. \\
\hline
Food (15 s)       & 1059 & 9747 & 59 & 226 & 18 \\
Food (30 s)       & 428  & 5775 & 59 & 213 & 28 \\
\hline
Cosmetics (15 s)  & 724  & 5861 & 55 & 181 & 20 \\
Cosmetics (30 s)  & 307  & 3851 & 54 & 177 & 34 \\
\hline
Car (15 s)        & 644  & 5954 & 58 & 204 & 19 \\
Car (30 s)        & 411  & 6913 & 55 & 197 & 30 \\
\hline
\end{tabular}
\vspace{0.5em}
\begin{flushleft}
\footnotesize
\item \textit{Notes}: “Vis.” and “Aud.” indicate the number of visual and audio expression-element types detected in the dataset, respectively. “Len.” denotes the maximum sequence length (number of scenes) used in the HMM-based modeling.
\end{flushleft}
\end{table}

\subsection{FE-Derived Structures in a Single Dataset}
For the single-dataset analysis, the 15 s food advertising dataset was used, which contains the largest number of samples among all categories.

\subsubsection{Expression Elements and Scene-Level FE Metrics}\label{sec:sceen_analysis}

\begin{figure}[t]
\centering

\includegraphics[width=0.8\columnwidth]{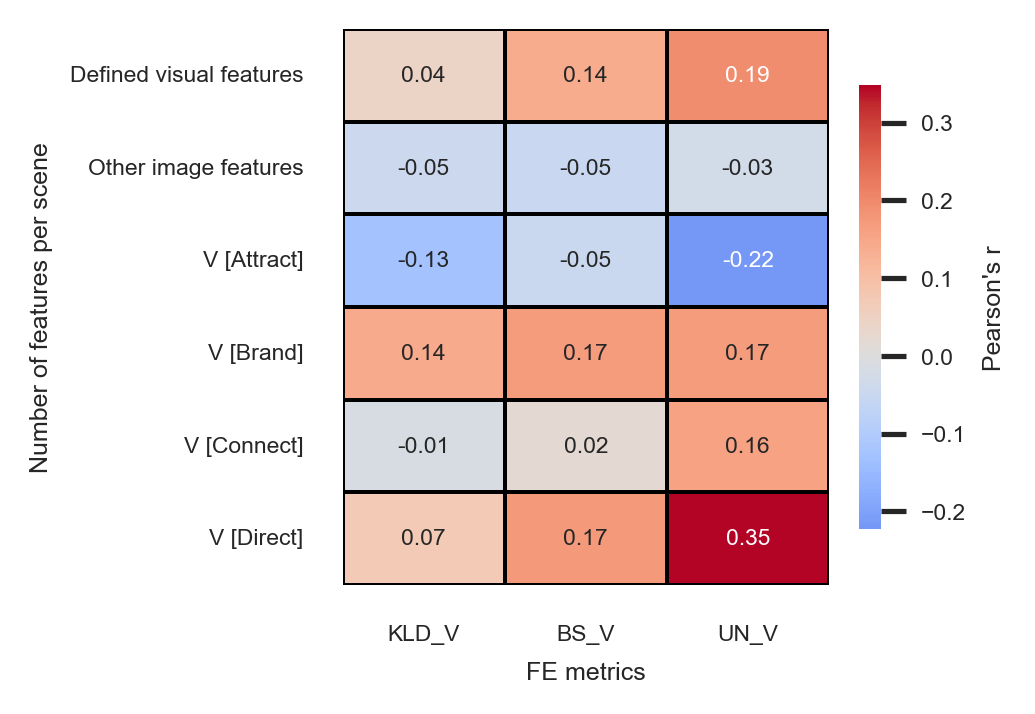}
\\[-2pt]
\makebox[0.8\columnwidth][l]{\footnotesize (a) Visual features vs. FE metrics.}

\includegraphics[width=0.8\columnwidth]{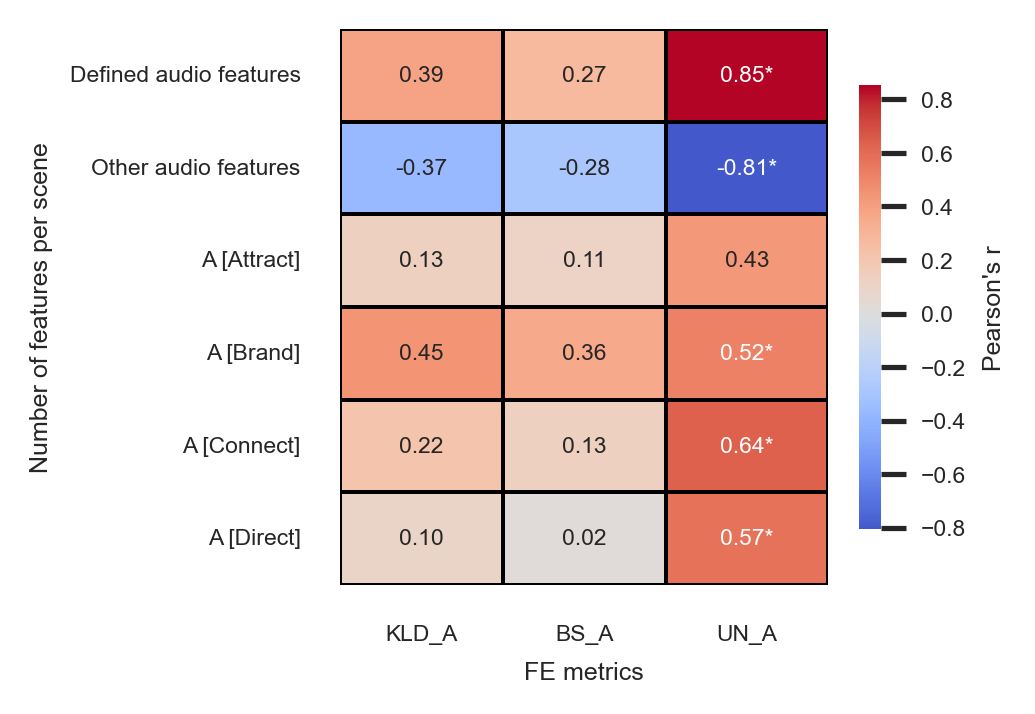}
\\[-2pt]
\makebox[0.8\columnwidth][l]{\footnotesize (b) Audio features vs. FE metrics.}
\caption{Correlation between FE metrics and advertising features in video scenes.}
\label{fig:scene_feature_corr}
\begin{flushleft}
\footnotesize
Parameters labeled with “V” or “A” represent expression features based on vision and audio, respectively, and they appear on both the x- and y-axes.
\end{flushleft}
\end{figure}

Scene-level analysis explored how FE metrics correspond to advertising expression characteristics. For each scene, occurrence counts of expression elements were aggregated, and Pearson’s correlation coefficients with significance probabilities were computed (Fig.~\ref{fig:scene_feature_corr}).

Expression elements were grouped into the four categories listed in Table~\ref{tab:ad_elements} (Attract, Brand, Connect, and Direct), and totals were calculated at the scene-level. Additional aggregate indices were created for visual and audio features, including elements outside predefined categories. FE metrics were computed by simulating the visual and audio modalities separately.

In the visual modality, no strong correlations emerged, yet several tendencies appeared:

\begin{itemize}
\item KLD: positive association with brand-oriented elements.
\item BS: positive associations with the total number of visual features and with brand-oriented or direct action–inducing elements.
\item UN: negative association with attention-attracting elements and positive association with the total number of visual features and direct action–inducing elements.
\end{itemize}

In the audio modality, clearer associations were observed:

\begin{itemize}
\item KLD: positive associations with the total number of audio features and with brand-oriented elements.
\item BS: positive associations with the total number of audio features and with brand-oriented elements.
\item UN: negative association with out-of-category audio elements, and positive associations with Connect elements, direct action–inducing elements, and the total number of audio expression features.
\end{itemize}

These findings show that visual and audio modalities present distinct but interpretable patterns of association with FE metrics.

\subsubsection{Patterns in Video-Level Emotional Indices}
\label{sec:movie_analysis}
Pearson’s correlation coefficients were computed to examine relationships among the video-level representative emotional indices (peak, end, skew, and decay rate)(Table~\ref{tab:corr_matrices}). The FE metrics were computed by integrating both the visual and audio modalities. The analysis revealed that the peak tended to show moderate correlations with other indices; however, all coefficients remained weak to moderate, with the strongest value observed at $|r| = 0.62$. These values fall below the commonly used threshold for strong multicollinearity ($|r| \ge 0.8$), confirming that the four indices were not redundant.

\begin{itemize}
\item For KLD and BS, moderate positive correlations were found between peak and skew ($r \approx 0.49$–$0.52$) and between peak and decay rate ($r \approx 0.27$–$0.51$).

\item For UN, a positive correlation appeared between peak and skew ($r = 0.32$), whereas the peak and decay rate showed the strongest correlation ($r = 0.62$, $p < 0.05$). 
\end{itemize}

These results confirm that, although some degree of association existed, each index captured a distinct aspect of emotional responses.

\begin{table}[t]
\centering
\caption{Correlation Matrices of FE Metrics (KLD, BS, and UN)}
\label{tab:corr_matrices}
\begin{tabular}{lcccc}
\hline
 & peak & end & Skew & Decay Rate \\
\hline
\multicolumn{5}{c}{\textbf{KLD}} \\
peak(KLD)        & 1.00* & 0.32 & 0.52* & 0.27 \\
end(KLD)         &        & 1.00* & -0.04 & 0.08 \\
Skew(KLD)        &        &        & 1.00* & 0.18 \\
Decay rate(KLD)   &        &        &        & 1.00* \\
\hline
\multicolumn{5}{c}{\textbf{BS}} \\
peak(BS)         & 1.00* & 0.30 & 0.49 & 0.51* \\
end(BS)          &        & 1.00* & -0.00 & 0.25 \\
Skew(BS)         &        &        & 1.00* & 0.16 \\
Decay rate(BS)    &        &        &        & 1.00* \\
\hline
\multicolumn{5}{c}{\textbf{UN}} \\
peak(UN)         & 1.00* & 0.27 & 0.32 & 0.62* \\
end(UN)          &        & 1.00* & -0.11 & 0.28 \\
Skew(UN)         &        &        & 1.00* & -0.11 \\
Decay rate(UN)    &        &        &        & 1.00* \\
\hline
\end{tabular}
\end{table}

Videos were clustered using the k-means method because the representative emotional indices retained independent information (Fig.~\ref{fig:food_cluster}). Cluster-number selection based on standard criteria (Appendix C~\ref{appendix:k_selection}) identified $k = 3$ as the most stable choice. The resulting clusters, namely C1 (n = 311), C2 (n = 359), and C3 (n = 389), showed distinct emotional patterns:

\begin{figure}[t]
\centering
\includegraphics[width=\columnwidth]{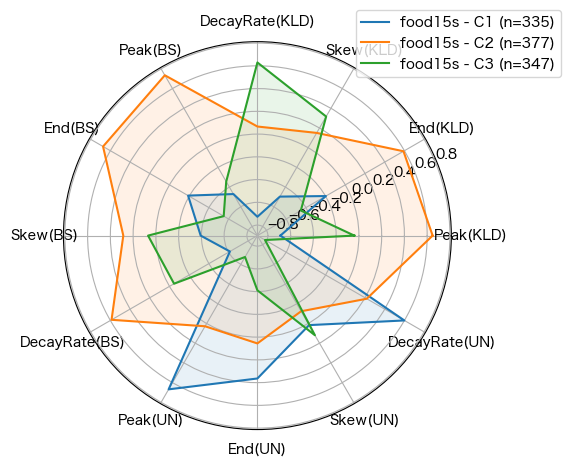}
\caption{Mean FE profiles for the food cluster structure}
\label{fig:food_cluster}
\end{figure}

\begin{itemize}
\item C1: consistently low BS and KLD, with a pronounced peak and relatively high decay rate in UN.
\item C2: uniformly high values across most indices, including the peaks and ends of BS, KLD, and UN.
\item C3: characterized by a markedly high decay rate of KLD.
\end{itemize}

These findings indicate that the proposed indices effectively grouped videos according to their emotional response patterns.

We examined whether the frequencies of advertising expression elements differed across the three clusters.
One-way ANOVA revealed significant differences for several variables, including visual attract elements ($F = 22.48$, $p < 10^{-9}$, $\eta^{2}=0.041$), audio attract elements ($F = 20.54$, $p < 10^{-8}$, $\eta^{2}=0.037$), 
audio brand elements ($F = 15.54$, $p < 10^{-6}$, $\eta^{2}=0.029$), and audio direct elements ($F = 14.29$, $p < 10^{-5}$, $\eta^{2}=0.026$). In contrast, visual direct and audio connect elements showed no clear cluster differences ($p > 0.05$). These results indicate that the clusters capture meaningful distinctions in expression features.

\subsubsection{Qualitative Analysis of Representative Samples}
\begin{figure*}[t]
\centering

\begin{minipage}{0.48\textwidth}
  \centering
  \includegraphics[width=\linewidth]{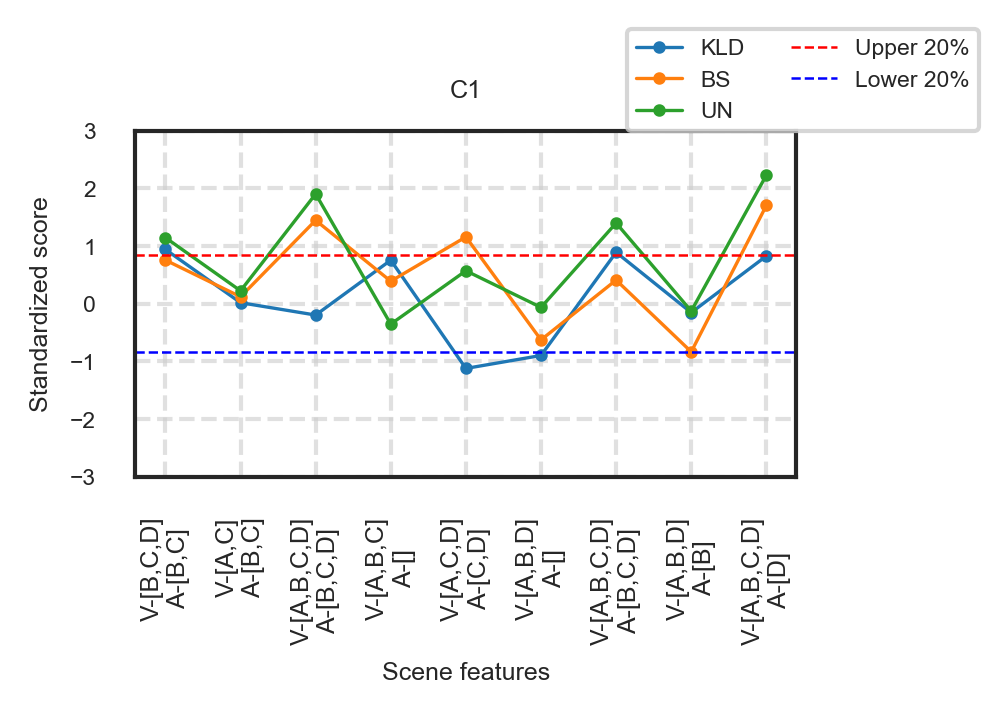}
  \vspace{2pt}
  \makebox[\linewidth][l]{\footnotesize (a) Cluster C1 example\par}
\end{minipage}\hfill
\begin{minipage}{0.48\textwidth}
  \centering
  \includegraphics[width=\linewidth]{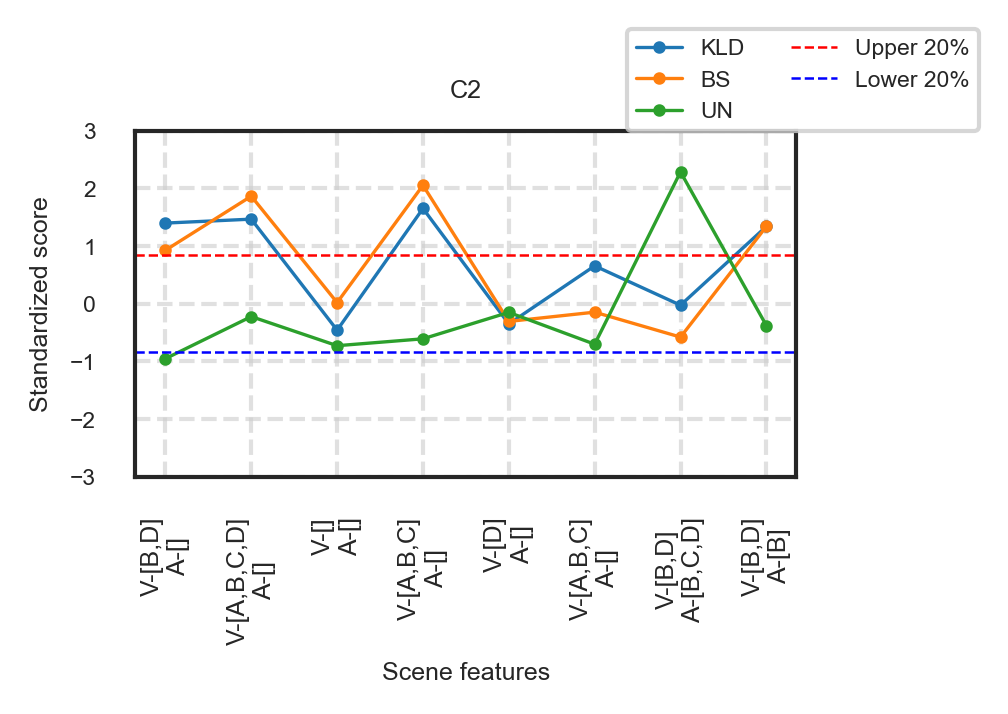}
  \vspace{2pt}
  \makebox[\linewidth][l]{\footnotesize (b) Cluster C2 example\par}
\end{minipage}

\vspace{8pt}

\begin{minipage}{0.48\textwidth}
  \centering
  \includegraphics[width=\linewidth]{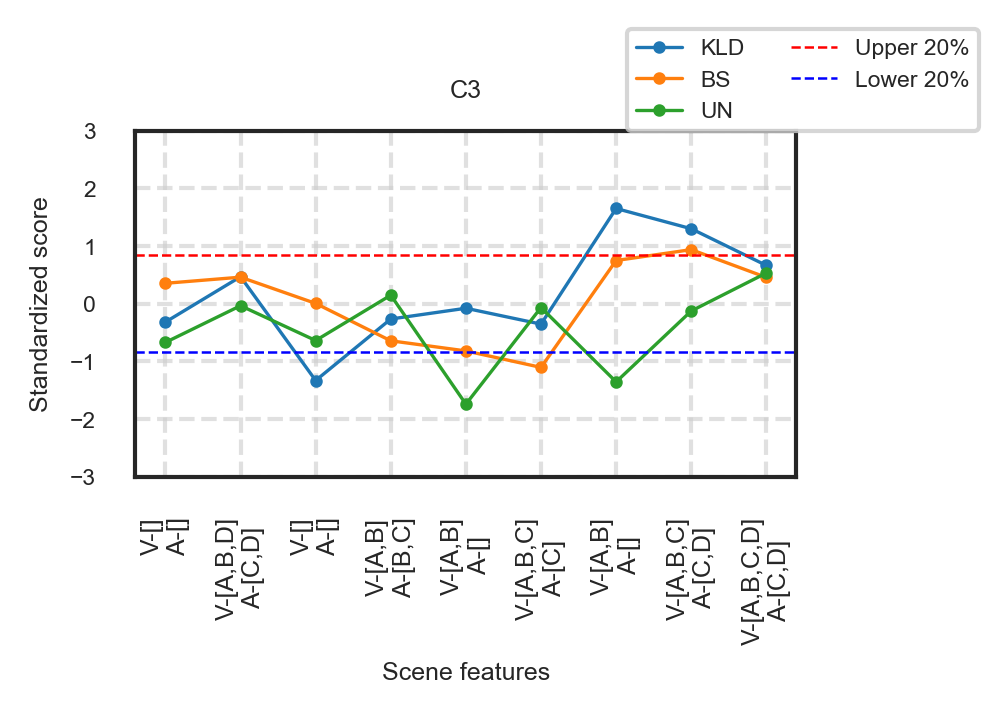}
  \vspace{2pt}
  \makebox[\linewidth][l]{\footnotesize (c) Cluster C3 example\par}
\end{minipage}\hfill

\caption{Representative example scenes and FE metrics from each cluster.}
\label{fig:cluster_example}
\begin{flushleft}
\footnotesize
X-axis: Combinations of scene features. “V” represents visual features and “A” represents audio features. The subscripts [A],[B],[C], and [D] indicate whether each of the four expression categories (Attract, Brand, Connect, and Direct) is included or excluded in the scene.
\end{flushleft}
\end{figure*}

One representative video was selected per cluster and scene trajectories of the advertising expression features and FE metrics were presented to illustrate the characteristics of each cluster. Representative samples were selected as those with the median number of scenes (nine) and a high cosine similarity to the centroid (mean values) of each cluster (Fig.~\ref{fig:cluster_example}).

\begin{itemize}
\item C1: UN remained consistently high, whereas KLD remained low. For advertising expression features, the presentation frequency of the components was high across the entire video in both visual and audio modalities.

\item C2: BS and KLD remained consistently high throughout the video, whereas UN remained low. Multiple composite visual elements (V[A, B, C] and A[]) characterized the increase in BS and KLD in advertising expression features.

\item C3: BS, KLD, and UN remained low throughout the video, with only slight increases in all metrics toward the end. For advertising expression features, both visual and audio elements were limited, although product and character close-ups (V[A]) appeared frequently.
\end{itemize}

These representative cases indicated that the proposed indices can effectively visualize the scene-level dynamics and differences in advertising expression features across clusters.

\subsubsection{Sensitivity Analysis of Model Hyperparameters}
This section evaluates whether the scene- and video-level FE structures remain robust to variations in the HMM hyperparameters. Sensitivity analyses were conducted around the optimized settings in Appendix B~\ref{appendix:hyperparam}, focusing on the number of hidden states and the Dirichlet prior scale.

\begin{figure}[t]
\centering

\includegraphics[width=0.8\columnwidth]{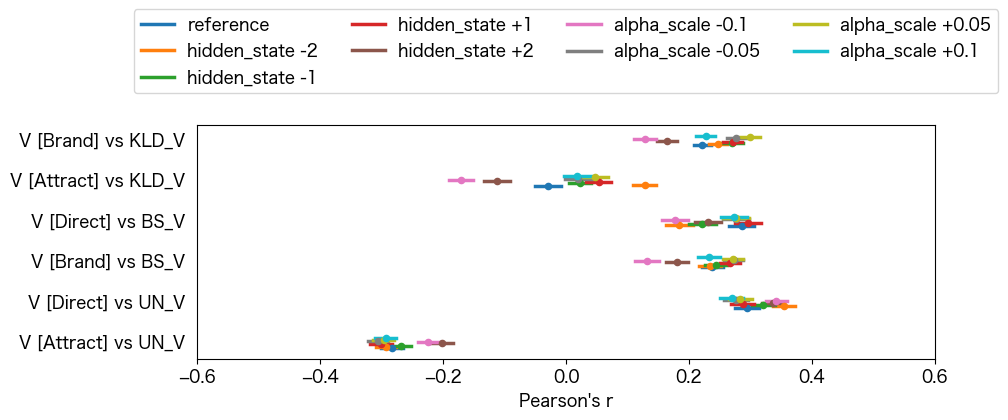}
\\[-2pt]
\makebox[0.8\columnwidth][l]{\footnotesize (a) Visual modality}

\includegraphics[width=0.8\columnwidth]{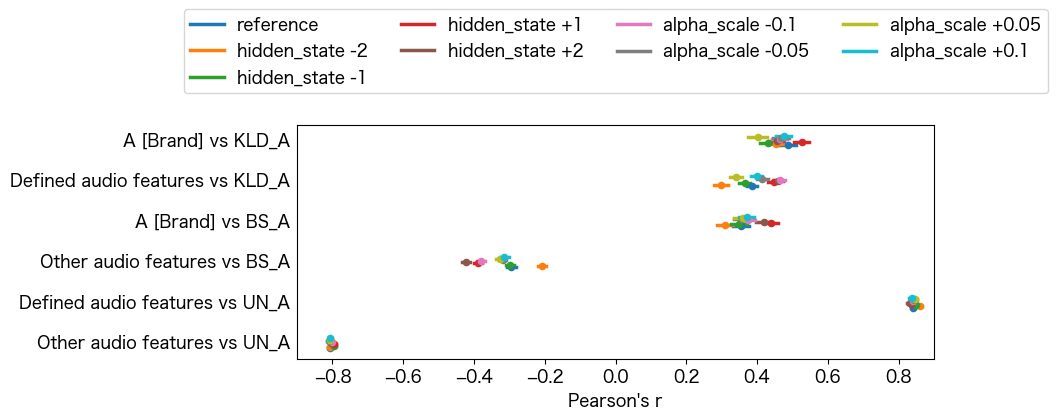}
\\[-2pt]
\makebox[0.8\columnwidth][l]{\footnotesize (b) Audio modality.}

\caption{Sensitivity analysis of visual and audio modalities.}
\label{fig:sensitivity_analysis}

\begin{flushleft}
\footnotesize
Each panel shows how variations in model parameters affect feature representations for the visual (a) and audio (b) modalities. The y-axis displays only the top two advertising expression features with the highest absolute correlations for each FE metric (KLD, BS, and UN).
\end{flushleft}
\end{figure}

For the scene-level, confidence intervals of the correlations were estimated using bootstrap sampling. As shown in Fig.~\ref{fig:sensitivity_analysis}, the bootstrap confidence intervals exhibited only minor fluctuations across hyperparameter settings, indicating that the correlation patterns were highly consistent. Only the visual attention–attracting element V[Attract] showed greater sensitivity to hyperparameter variation, exhibiting larger fluctuations compared with the other elements. These results indicate that the scene-level FE–metrics relationships remain largely unchanged around the optimized hyperparameter settings, despite variations in the number of hidden states (-2, -1, +1, and +2) and the Dirichlet prior scale (-0.10, -0.05, +0.05, and +0.10).

To assess how each hyperparameter affected the reproducibility of the clustering results, the Adjusted Rand Index (ARI) was computed between the reference model and each perturbed setting. For the hidden state variations, the ARI values ranged from 0.24–0.43, while for the alpha scale variations, the values ranged from 0.20–0.62. These results indicate that the clustering structure remained moderately stable across a broad range of parameter changes.

\subsection{Genre and Duration Generalizability}
\begin{figure}[t]
\centering

\includegraphics[width=0.8\columnwidth]{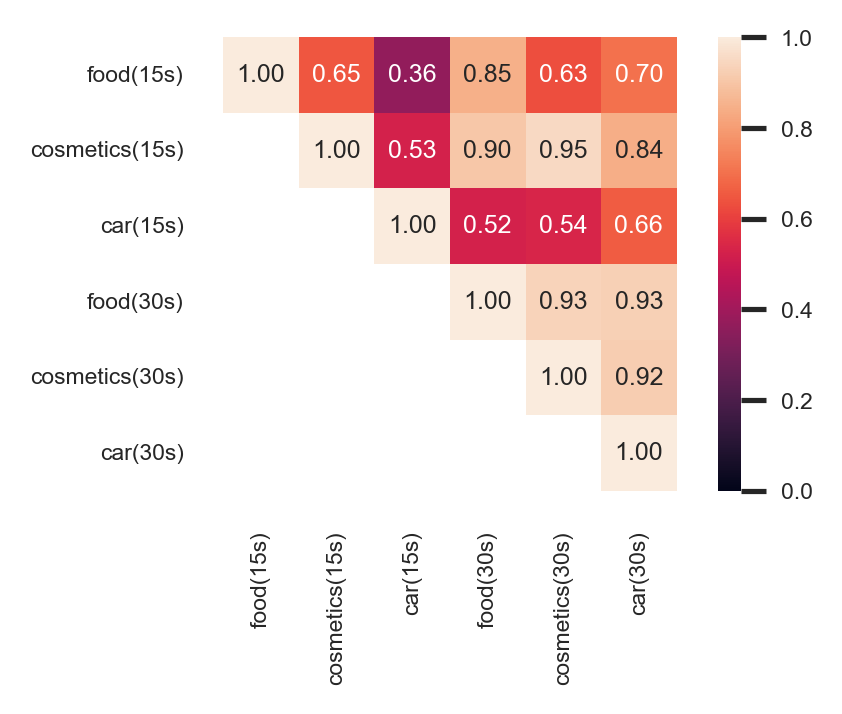}
\\[-2pt]
\makebox[0.8\columnwidth][l]{\footnotesize (a) Visual modality}

\includegraphics[width=0.8\columnwidth]{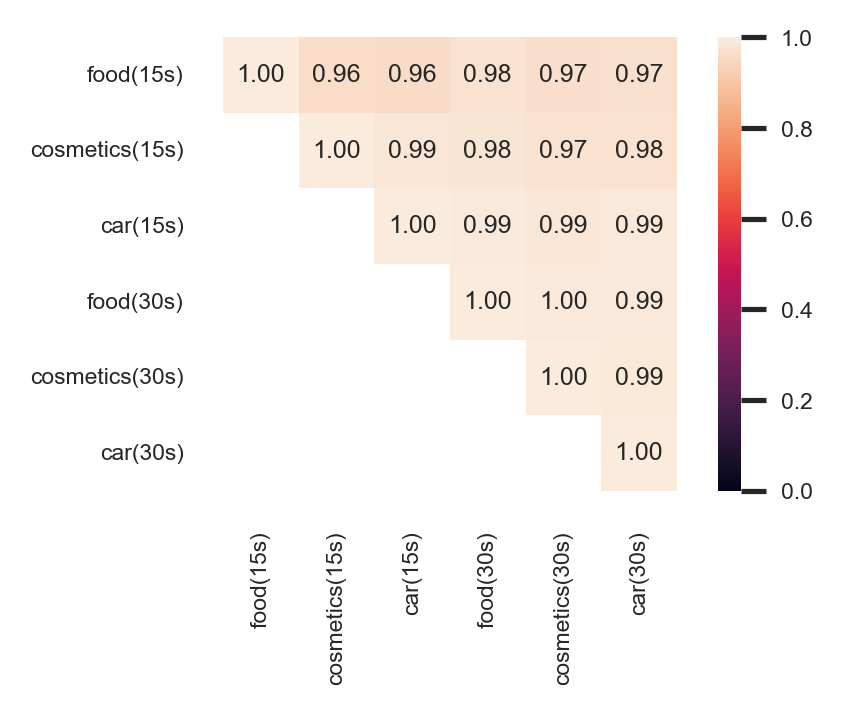}
\\[-2pt]
\makebox[0.8\columnwidth][l]{\footnotesize (b) Audio modality.}

\caption{Cross-dataset matrix correlations of scene-level FE structures}
\label{fig:cross_condition_corr}

\begin{flushleft}
\footnotesize
\end{flushleft}
\end{figure}

In this section, HMMs were trained for datasets differing in genre and video duration, and the corresponding FE metrics were computed.
For the scene-level associations between advertising expressions and FE metrics, each correlation matrix was vectorized, and matrix correlations were calculated across datasets (Fig.~\ref{fig:cross_condition_corr}).

For audio-based expressions, the matrix correlations were close to 1.0 across all combinations of genres and video durations, indicating that the same relational structure emerged consistently.
For visual expressions, the lowest correlation was 0.36, and all combinations yielded similarity above a certain level. Higher similarity appeared within the same genre or the same video duration, while relatively lower correlations were observed between the car genre and the other genres. Differences in representative correlation patterns across genres are provided in Appendix D.

\begin{figure}[t]
\centering
\includegraphics[width=\columnwidth]{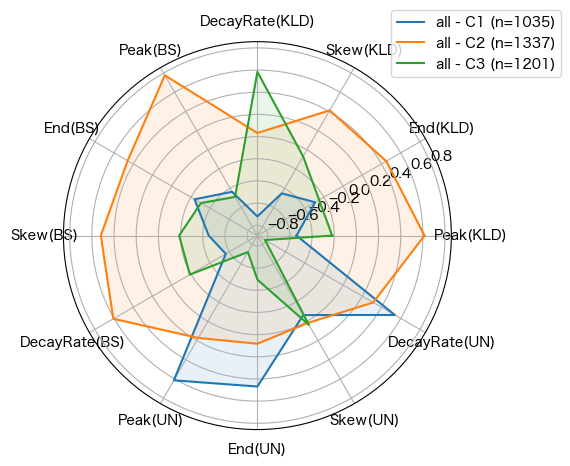}
\caption{Mean FE profiles for the common cluster structure}
\label{fig:common_cluster}
\end{figure}

The video-level FE metrics from all datasets were then combined for clustering, and the mean FE profile for each cluster was derived (Fig.~\ref{fig:common_cluster}).
The resulting cluster structures showed strong similarity to the reference cluster structure obtained from the 15-second food genre (Fig.~\ref{fig:food_cluster}).

Finally, the chi-square test and standardized residuals used to assess genre differences in cluster distribution are presented in the Appendix D.

\section{Discussion}
This study interpreted the KLD as an index of pleasantness and regarded BS and UN as indices of surprise. In addition, habituation was interpreted as the reduction in these indices under repeated exposure.

\subsection{Usefulness of Scene-Level FE Metrics} The scene-level trends in Fig.~\ref{fig:scene_feature_corr} demonstrate systematic relationships between FE metrics and advertising expression elements. Across both modalities, KLD exhibited positive associations with brand-oriented elements, indicating that the “pleasantness” index consistently responds to brand presentation regardless of modality.

BS exhibited positive associations with brand-oriented elements in both the visual and audio modalities, and in the visual modality, it was also associated with the total number of features and with direct action–inducing elements. These tendencies suggest that the learning-driven component of “surprise” is sensitive to informational complexity, particularly the multiplicity and co-occurrence of elements.

UN revealed modality-specific patterns. In the visual modality, UN showed a negative association with attention-attracting elements and a positive association with direct action–inducing elements. In contrast, in the audio modality, UN demonstrated positive associations with Connect elements, direct action–inducing elements, brand-oriented elements, and with the total number of audio expression features, while showing a negative association with out-of-category audio elements. These results indicate that the uncertainty-based “surprise” index responds differently depending on the modality and is sensitive to both the type and quantity of presented elements.

Taken together, these findings suggest that FE metrics tend to reflect the structural properties of advertising expression elements. This indicates that the proposed metrics may be useful for characterizing emotional responses at the scene level and may capture modality-dependent variations in informational structure.

\subsection{Usefulness of Video-Level Emotional Indices}
The correlation analysis of the video-level representative indices (Table~\ref{tab:corr_matrices}) showed that correlations remained weak to moderate. The absence of strong multicollinearity confirms that each index captured a distinct dimension of emotional responses.

The consistent associations between the peak and skew, as well as between the peak and decay rate across the KLD, BS, and UN, suggest that emotional peaks reflect both the distributional characteristics of responses and habituation-related dynamics. In contrast, the weak correlations of the end index with the other indices indicate that emotional impressions formed at the conclusion of a video provide complementary information that is not explained by peak- or learning-related patterns.

The usefulness of the proposed indices was further supported by the k-means clustering results (Fig.~\ref{fig:food_cluster}). The three clusters identified in the results section can be interpreted as C1: Uncertain Stimulus, characterized by low BS and KLD together with a pronounced UN peak; C2: Sustained High Emotion, showing uniformly high responses across all indices; and C3: Momentary Peak and Decay, characterized by a high decay rate in the KLD. These clusters demonstrate that the indices differentiate videos with distinct emotional structures. The same differences were also evident in the scene-level trajectories of representative videos in each cluster (Fig.~\ref{fig:cluster_example}), confirming that the responses of the KLD, BS, and UN align with the scene-level correlation patterns.

These cluster-level distinctions closely correspond to the representative indices and offer interpretable patterns of emotional responses in video advertisements. Taken together, the observed correlation structure (Table~\ref{tab:corr_matrices}) and the degree of cluster separation (Fig.~\ref{fig:food_cluster}) suggest that the indices of “pleasantness,” “surprise,” and “habituation” play complementary roles. This indicates that the proposed indices may serve as a useful basis for organizing video advertisements according to multiple emotional profiles.


\subsection{Robustness Across Model Parameters}
Sensitivity analyses showed that even substantial changes in the HMM hyperparameters did not disrupt the FE-based structures at either the scene- or video-level. As shown in Fig.~\ref{fig:sensitivity_analysis}, the bootstrap confidence intervals for the scene-level correlations were extremely narrow, indicating that variations in the number of hidden states or the scale of the Dirichlet prior produced minimal influence on the relationships between expression elements and FE metrics. This stability suggests that the FE metrics capture structural properties intrinsic to advertising content, including food-related videos.

At the video-level, the cluster structures also remained largely robust to parameter modifications, with ARI values ranging from 0.20–0.62, indicating moderate reproducibility. Although some variation appeared in the specific cluster assignments, the main structural patterns were preserved. This result implies that the emotional profiles of videos arise from stable differences in the representative emotional indices rather than from fine-grained model settings.

Taken together, the FE-based framework consistently extracts the underlying emotional structure of advertising videos without depending on a particular hyperparameter configuration. Such robustness supports the use of these emotional metrics as stable indicators for practical advertising analysis and creative evaluation.

\subsection{Cross-Condition Generalizability}
The datasets exhibited clear differences across categories and durations (Table~\ref{tab:dataset_stats}). For example, the relative emphasis on visual versus auditory elements varied depending on whether the advertisement centered on products or human characters, and shorter durations tended to compress information, increasing the diversity of expressive patterns. Given these differences, this study examined the extent to which the structures of the FE metrics were preserved across conditions that differed in genre and video duration. The analyses showed that audio expressions exhibited extremely high similarity across all datasets(Fig.\ref{fig:cross_condition_corr}), suggesting that auditory stimuli play a relatively consistent role in advertising. In contrast, visual expressions demonstrated generally high similarity but also revealed genre-specific differences (see Appendix D for details), likely reflecting distinct visual presentation strategies across product categories.

Clustering based on the video-level FE metrics showed highly similar structures between all datasets (Fig.~\ref{fig:common_cluster}) and the 15 s food reference dataset (Fig.~\ref{fig:food_cluster}). This similarity indicates that the temporal patterns of KLD, BS, and UN captured abstract emotional responses that were not tied to specific advertising content.

Overall, the findings indicate that the FE metrics offer a flexible and generalizable representation: (i) they maintain relatively stable structures across genres and video durations, and (ii) they can reflect genre-specific differences when present.

\subsection{Overall Implications}
The experimental findings have several overall implications.

First, because the scene-level FE metrics revealed consistent structural relationships with visual and auditory advertising expression elements, the proposed metrics provide a quantitative basis for interpreting scene-level emotional effects.

Second, as the video-level emotional indices enabled the classification of advertisements into distinct characteristic patterns, they offer a practical framework for organizing large-scale video datasets according to emotional structure.

Third, the robustness of the FE structures to variations in HMM hyperparameters indicates that the proposed method is stable and reliable across modeling choices.

Finally, the generalizability of FE structures across genres and video durations suggests that the proposed framework captures fundamental emotional response patterns that are not limited to specific advertising categories or formats.

Taken together, these implications suggest that the proposed FE-based indices provide a coherent and interpretable basis for quantitatively examining emotional responses to video advertisements, with the potential to be applied in scalable analyses.

\subsection{Limitations and Future Directions}
This study has some limitations arising from its scope and methodological design, and clarifying them is important for identifying directions for future research.

\subsubsection{Defining and Extracting Expression Elements}
The analysis did not explicitly incorporate higher-order temporal structures, such as narrative flow, and the accuracy of expression-element extraction remains limited. Classifications produced by VLMs and LLMs depend on prompt design, although they offer the advantage of capturing elements that creators consider important. Future work may benefit from integrating advanced feature extraction methods, such as multi-modal transformers or diffusion models, with FE computation to analyze richer expressive structures.

\subsubsection{Modality Interaction and Model Extensions}
The present model does not explicitly address interactions or weightings between visual and auditory modalities. Future studies will require mechanisms that adaptively learn modality weights, as well as more flexible model structures, including hierarchical HMMs and deep neural generative models. By explicitly modeling how each modality contributes to the inferred emotional states, such adaptive weighting mechanisms are expected to enhance the interpretability of the model and clarify which visual or auditory cues drive changes in FE metrics.

\subsubsection{Comparison With Existing Emotion-Estimation Models}
This study did not compare model performance with existing emotion-estimation methods, such as convolutional neural network (CNN)- or recurrent neural network (RNN)-based models trained on subjective emotional responses to visual or auditory stimuli. Establishing appropriate data acquisition procedures will be necessary to clarify the positioning and effectiveness of the proposed approach.

\subsubsection{Dataset and Generalizability Considerations}
The analysis was limited to TV advertisements in Japan. To evaluate broader generalizability, future work should incorporate advertising datasets from other cultural contexts (e.g., Western markets), additional product categories, advertisements of different lengths, and variations in editing tempo, such as the scene-change speed and auditory rhythm. The tempo and rhythm may strongly influence emotional dynamics, highlighting the need to extend temporal feature representations.

\subsubsection{FE Metrics and Subjective or Behavioral Outcomes}
The correspondence between FE metrics and subjective emotional ratings, dropout rates, brand memory, or other advertising outcomes has not yet been examined. To strengthen the interpretability and external validity of these FE metrics, future experiments should collect subjective emotional ratings that clarify how KLD, BS, and UN correspond to self-reported pleasantness, surprise, and habituation. Furthermore, examining how the relationships between FE metrics and subjective emotions differ by age or gender will help clarify variations in emotional responses across demographic groups.

Establishing psychological and practical validity will also require integrating additional behavioral information, including gaze data, long-term viewing histories, and other behavioral logs. For example, the habituation-related decay rate provides a quantitative indicator of how rapidly emotional responses diminish with repeated exposure, enabling the identification of the point at which additional ad impressions generate diminishing emotional or mnemonic returns. The combination of KLD peaks and habituation-related decay rates can also characterize responses that may be described as “pleasant yet quickly habituating,” which has potential applications for estimating the optimal frequency of advertising exposure.

Building on prior research linking emotional responses to brand memory and user retention, the emotional dynamic indicators estimated in this study could further support analyses of how individual viewing histories and emotional trajectories relate to advertising effectiveness. For instance, sustained emotional engagement may help maintain attention and reduce dropout, whereas brief but intense emotional responses may contribute more strongly to brand memory and attitude formation. In addition, interest-like surprise may help prevent user disengagement, while pleasantness may facilitate the consolidation of brand memory. Taken together, these emotional dynamic patterns provide a quantitative foundation for future investigations into how specific forms of emotional trajectory contribute to user retention and brand memory.

\subsubsection{Application to AI-Generated Content}
Artificial intelligence (AI)-generated advertisements may exhibit limited variability or emotional nuance when explicit prompt control is insufficient. Applying the proposed method could help researchers to diagnose the emotional properties of AI-generated content in advance. Furthermore, incorporating FE metrics as evaluation functions within generative models such as diffusion models may support the creation of advertisements designed to elicit specific emotional trajectories, including momentary surprise or sustained pleasantness. In addition, modeling emotional dynamics in human–computer interaction videos, such as AI anchors, represents a promising direction for understanding how viewers respond to algorithmically generated communicative behavior. Applications to human–AI interaction videos, more broadly, also appear promising.

\section{Conclusion}
This study introduced a method for mathematically quantifying “pleasantness,” “surprise,” and “habituation” in videos based on the FEP and examined its usefulness using a large-scale dataset of advertising videos.

At the scene-level, the proposed indices showed systematic correspondence with advertising expression elements, suggesting that they provide a quantitative basis for interpreting momentary emotional effects. At the video-level, representative indices such as peak, end, skew, and decay rates retained independent information, indicating that they may be useful for characterizing overall emotional responses from multiple perspectives. Cluster analysis further showed that the indices helped organize large sets of videos into distinct emotional patterns, including Uncertain Stimulus, Sustained High Emotion, and Momentary Peak and Decay.

Robustness analyses suggested that the FE-based structures remained stable across variations in HMM hyperparameters, and generalization tests across genres and video durations indicated that the framework may capture broader emotional patterns that are not specific to advertising categories.

Taken together, these findings suggest that the proposed FE-based framework offers a coherent and interpretable basis for quantitatively estimating emotional responses to video advertisements solely from their expression features, without relying on subjective or physiological data. Future work should further extend the applicability of this method by incorporating a wider range of expression elements and validating the framework through subjective viewer evaluations across different product categories and cultural contexts. Such efforts may enhance the methodological robustness and generalizability of the proposed approach.

\appendix
\section*{Appendix A: Prompt Design Details}\label{appendix:prompt}
\renewcommand{\lstlistingname}{Prompt}
\renewcommand{\thelstlisting}{\arabic{lstlisting}}
\setcounter{lstlisting}{0}

Prompt~\ref{fig:A1} shows the instruction used with the VLM to annotate expression elements in images from video advertisements, and Prompt~\ref{fig:A2} shows the instruction used with the LLM to annotate expression elements in utterances. The latter was implemented in Japanese, with metadata, such as the company name, product name, and brand name, inserted for each video. Regarding the extraction accuracy, the image-based prompt (Prompt~\ref{fig:A1}) produced unstable outputs when the instruction text became long, and extracting named entities from images was generally more difficult than with the text-based prompt. In addition, due to data transmission constraints, both the LLM and VLM were run locally using the most accurate small-scale publicly available models that could be executed within the in-house environment.
\begin{lstlisting}[caption={Prompt for annotating images in video ad}, label={fig:A1}, captionpos=b, abovecaptionskip=6pt]
Please answer "yes" or "no" for each of the following advertising elements, one at a time:
# Advertising Elements
    1. An image of the company logo mark
    2. An image of the food product
    3. An image of people or characters
    4. A close-up shot of the food product
    5. A close-up shot of people or characters
    6. Text describing the food product
    7. Text encouraging viewers to buy
    8. Text motivating viewers to consider buying
# Format
- [Number]: "yes" or "no"
\end{lstlisting}

\begin{lstlisting}[caption={Prompt for annotating utterances in video ad (food dataset)}, label={fig:A2}, captionpos=b, abovecaptionskip=6pt]
Please answer "yes" or "no" for each of the following advertising elements, one at a time, with respect to the narration or the utterances of characters in each scene of a food-related video advertisement.Your answers should be based on the factual content included in the utterance.  
# Advertising Elements
    1. A proper noun or general term close to the brand name "{brand_name}"  
    2. A proper noun or general term close to the product name "{product_name}"  
    3. A proper noun close to the company name "{company_name}"  
    4. Product description: a sentence that specifically describes the features of the product  
    5. Purchase promotion: a sentence that encourages viewers to buy the product  
    6. Purchase motivation: a sentence that presents reasons or motivations for purchasing the product  
    7. Direct address: a sentence that speaks directly to the viewer  
    8. Positive wording: a sentence containing expressions of positive feelings or values  
    9. Catchphrase: a short expression emphasizing the appeal or superiority of the product/service  
       (e.g., through repetition, comparison, emotional wording, numbers, future orientation,
       emphatic endings, or imperative forms)  

# Utterance in each scene
{Utterances}

# Output format
Please output each utterance in one line.
The format should be:
Utterance ID: utterance content, yes/no x 9

Example:
Utterance ID: Utterance content, yes, no, yes, no, yes, no, yes, no, yes
\end{lstlisting}

\section*{Appendix B: Hyperparameter Settings and Search}\label{appendix:hyperparam}
This appendix summarizes the hyperparameter settings and search procedure used for training the variational Bayesian HMM.
Three hyperparameters were examined: (1) the number of hidden states, (2) the mini-batch learning rate, and (3) the scale parameter of the Dirichlet priors on $\mathbf{A}^{1}$, $\mathbf{A}^{2}$, $\mathbf{B}$, and $\mathbf{D}$.

The baseline setting used five hidden states, a Dirichlet scale of 1.0, and a mini-batch learning rate of 0.01, and each hyperparameter was varied independently while keeping the remaining ones fixed.

Hyperparameter search was conducted using the 15 s food dataset, which contained the largest number of samples among all categories. Models were trained using variational inference implemented in Pyro. The dataset was split 80:20 into training and validation sets, and the validation objective was the negative ELBO. For each hyperparameter configuration, the model was trained five times with different random seeds, and the mean validation loss with ±1 standard deviation across the five runs was recorded.

\subsection*{B.1 Number of Hidden States}

\begin{figure}[htbp]
    \centering
    \includegraphics[width=0.85\linewidth]{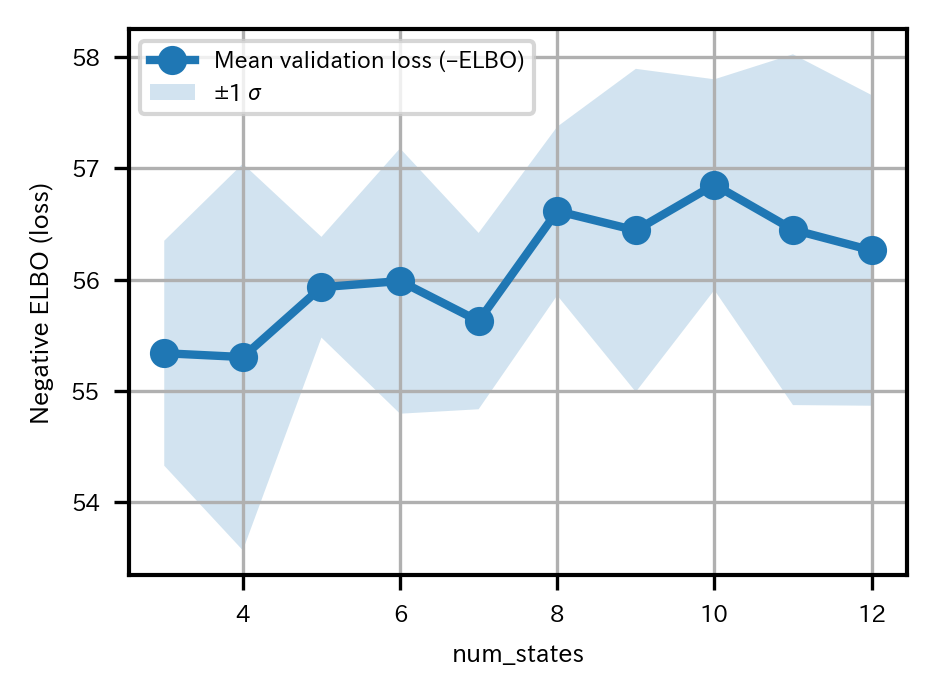}
    \caption{Validation loss vs.\ hidden state numbers ($\pm 1$ SD).}
    \label{fig:num_states_search}
\end{figure}

Figure~\ref{fig:num_states_search} shows the sensitivity of the validation loss with respect to the number of hidden states.
The configuration with five hidden states exhibited a relatively low mean validation loss with a small standard deviation, and was therefore adopted as the final setting.

\subsection*{B.2 Mini-Batch Learning Rate}

\begin{figure}[htbp]
    \centering
    \includegraphics[width=0.85\linewidth]{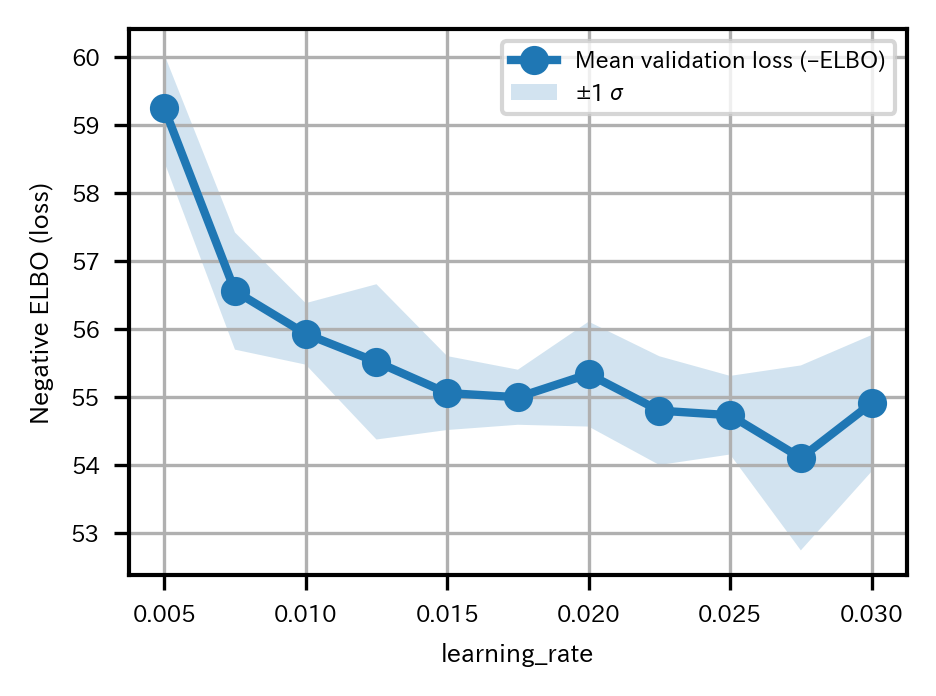}
    \caption{Validation loss vs.\ learning rates ($\pm 1$ SD).}
    \label{fig:lr_search}
\end{figure}

The impact of the mini-batch learning rate is summarized in 
Figure~\ref{fig:lr_search}.  
Learning rates between 0.02 and 0.03 produced stable convergence, 
and the optimal value selected based on the validation loss 
was $0.0275$.

\subsection*{B.3 Dirichlet Prior Scale Parameter}

\begin{figure}[htbp]
    \centering
    \includegraphics[width=0.85\linewidth]{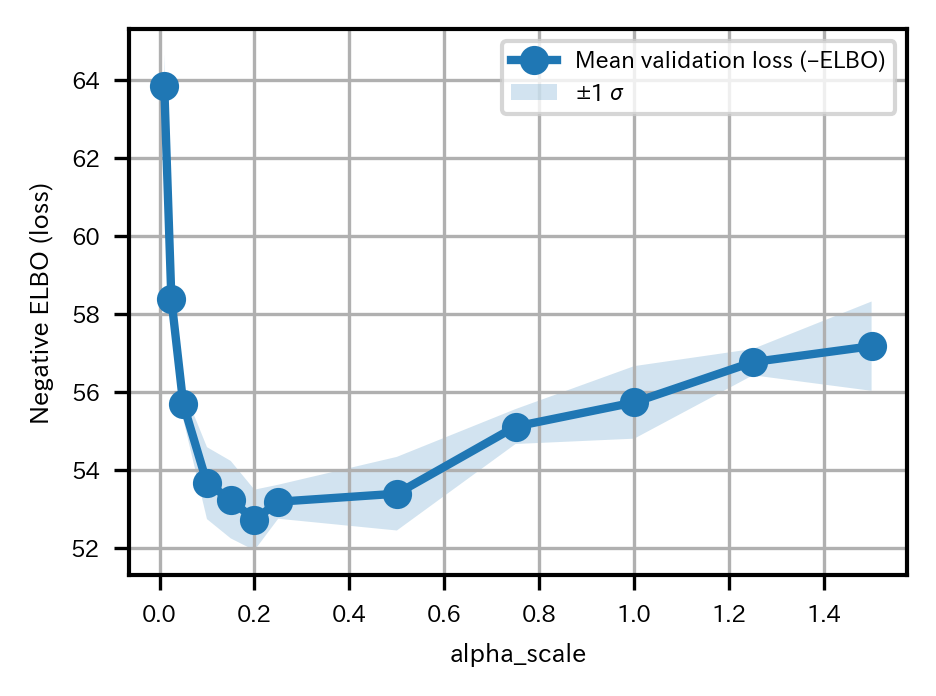}
    \caption{Validation loss vs.\ Dirichlet scales ($\pm 1$ SD).}
    \label{fig:alpha_scale_search}
\end{figure}

Figure~\ref{fig:alpha_scale_search} shows the effect of scaling the 
Dirichlet prior concentration on the model parameters.  
A scale value of $0.2$ consistently yielded the lowest validation loss 
and was chosen as the final configuration.

\subsection*{B.4 Final Hyperparameter Configuration}
These values were used consistently across all datasets for reproducibility.

\section*{Appendix C: Cluster Number Selection}\label{appendix:k_selection}
This appendix summarizes the quantitative criteria used to determine the number of clusters $k$. For each $k$, the silhouette score, the mean ARI across repeated runs, and the Kneedle distance calculated from the WCSS curve were computed. As listed in Table~\ref{tab:k_selection}, the candidate values were narrowed to $k=3$ and $k=4$, with $k=3$ showing the highest ARI and therefore providing the most stable solution.

\begin{table}[t]
\centering
\caption{Cluster Selection Metrics for $k=2$–10}
\label{tab:k_selection}
\begin{tabular}{c|ccc}
\hline
$k$ & Silhouette & Mean ARI & Kneedle Dist \\
\hline
2  & 0.12 & 0.56 & 0.00 \\
3  & 0.11 & 0.69 & 1.54 \\
4  & 0.10 & 0.40 & 1.97 \\
5  & 0.10 & 0.46 & 2.26 \\
6  & 0.10 & 0.48 & 2.16 \\
7  & 0.09 & 0.38 & 1.62 \\
8  & 0.09 & 0.34 & 1.31 \\
9  & 0.09 & 0.34 & 0.73 \\
10 & 0.08 & 0.30 & 0.00 \\
\hline
\end{tabular}
\end{table}

\section*{Appendix D: Cross-Condition Differences}\label{appendix:cond_differences}
This appendix provides supplementary results on differences in correlation structures and cluster distributions across datasets that vary in genre and video duration.

\begin{figure}[t]
\centering

\includegraphics[width=0.8\columnwidth]{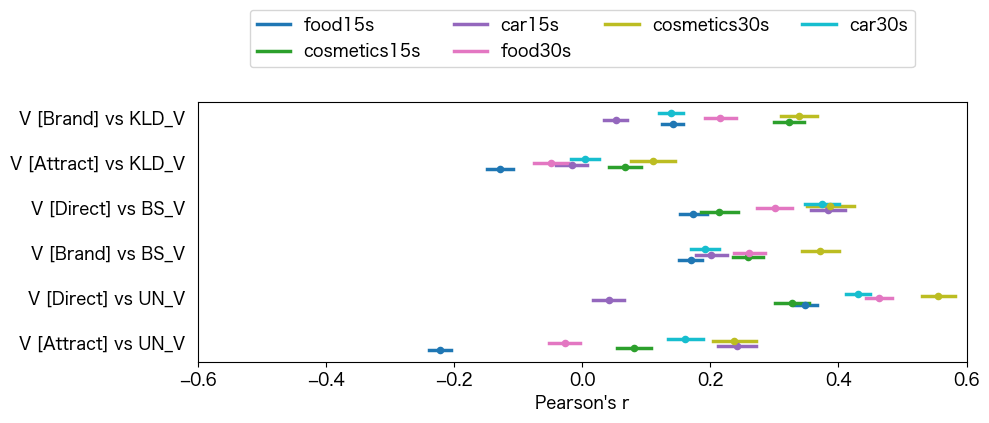}
\\[-2pt]
\makebox[0.8\columnwidth][l]{\footnotesize (a) Visual modality}

\includegraphics[width=0.8\columnwidth]{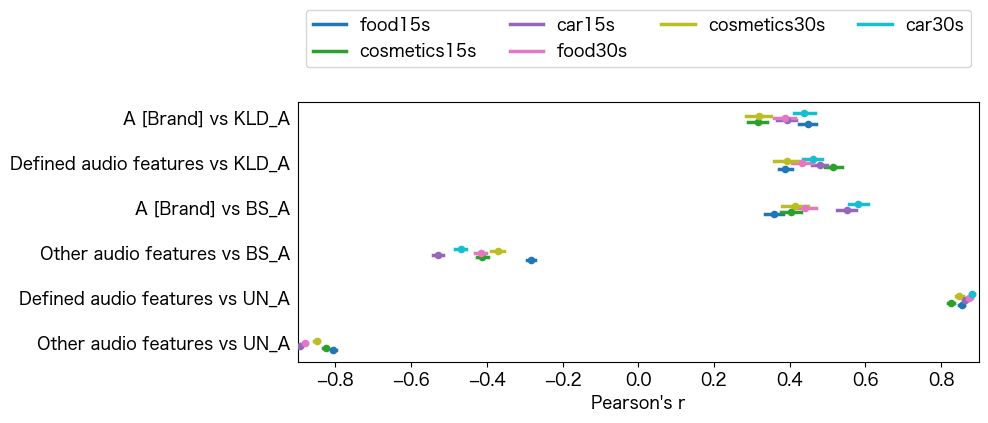}
\\[-2pt]
\makebox[0.8\columnwidth][l]{\footnotesize (b) Audio modality.}

\caption{Cross-dataset correlation comparison}
\label{fig:cross_dataset_r}

\begin{flushleft}
\footnotesize
\end{flushleft}
\end{figure}

First, representative correlations between advertising expressions and FE metrics were compared across datasets using bootstrap sampling, following the procedure used in the sensitivity analysis (Fig.~\ref{fig:cross_dataset_r}). This evaluation clarified how the correlation structures varied with changes in genre and video length. 

For visual expressions, some variability appeared across datasets, with the car genre showing notably lower correlations for direct-purchase cues (V[Direct]), possibly due to limited text detection on product packaging. In contrast, audio expressions showed a highly consistent correlation structure across all genres and video durations.

Next, a chi-square test was conducted to examine whether the cluster distribution from the integrated clustering analysis differed across genres. The test indicated a significant difference, $\chi^{2}(10)=341.41, p<.001$. To identify which genre–cluster combinations contributed to this result, standardized residuals were computed for each cell (Table~\ref{tab:std_residuals}). 

The frequency of C2, which reflects the highest emotional responses, varied substantially across genres and video durations. C2 appeared more frequently in 30 s videos and was relatively less common in 15 s videos. This tendency was particularly pronounced in the cosmetics and food genres, whereas the car genre showed comparatively weaker C2 responses.

\begin{table}[t]
\centering
\caption{Standardized Residuals by Condition and Cluster}
\begin{tabular}{lccc}
\hline
Genre & C1 & C2 & C3 \\
\hline
food(15s)          &  2.47 & -6.19 &  4.24 \\
food(30s)      & -1.61 &  6.86 & -5.74 \\
\hline
cosmetics(15s)     &  2.92 & -5.52 &  3.12 \\
cosmetics(30s)  & -3.81 &  7.57 & -4.45 \\
\hline
car(15s)           & 0.11 & -1.54 &  1.53 \\
car(30s)        & -3.03 &  5.66 & -3.16 \\
\hline
\end{tabular}
\label{tab:std_residuals}
\end{table}

\section*{Acknowledgment}
We would like to thank Editage (www.editage.jp) for English language editing. We are also grateful to M Data Co., Ltd. and PROJECT INC. for providing the video advertisement dataset and for their generous support.

\section*{Conflict of Interest}
The authors declare no competing interests.

\section*{Author Contributions}
TU conceived the study, designed the model, implemented the system, and conducted the experiments. TU also drafted the initial manuscript and revised the text throughout. HY supervised and guided the research, contributed to manuscript revision, and read and approved the final version. KO provided valuable advice during the manuscript revision in the peer-review phase. All authors are responsible for the content of the manuscript.

\section*{Data Availability Statement}
The data generated and analyzed in this study are considered confidential within the authors’ affiliated organization and cannot be made publicly available.

\bibliographystyle{IEEEtran}   
\bibliography{ref}            


\EOD

\end{document}